\def\year{2019}\relax
\documentclass[letterpaper]{article} 
\usepackage{aaai19}  
\usepackage{times}  
\usepackage{helvet} 
\usepackage{courier}  
\usepackage[hyphens]{url}  
\usepackage{graphicx} 
\usepackage{graphicx}  
\usepackage[vlined,ruled,linesnumbered]{algorithm2e}
\usepackage{amsfonts}
\usepackage{amssymb}
\usepackage{amsmath}
\newcommand{\omitit}[1]{}

\begin{document}

\title{Playing to Learn Better: Repeated Games for Adversarial Learning with Multiple Classifiers}
\author{Prithviraj Dasgupta, Joseph B. Collins, Michael McCarrick\\
Distributed Intelligent Systems Section (Code 5583)\\
Info. Mgmt. \& Decision Arch. (IMDA) Branch\\
Information Technology Division\\
U. S. Naval Research Laboratory, Washington, D.C.
}
\maketitle

\begin{abstract}
We consider the problem of prediction by a machine learning algorithm, called learner, within an adversarial learning setting. The learner's task is to correctly predict the class of data passed to it as a query. However, along with queries containing clean data, the learner could also receive malicious or adversarial queries from an adversary. The objective of the adversary is to evade the learner's prediction mechanism by sending adversarial queries that result in erroneous class prediction by the learner, while the learner's objective is to reduce the incorrect prediction of these adversarial queries without degrading the prediction quality of clean queries. We propose a game theory-based technique called a Repeated Bayesian Sequential Game where the learner interacts repeatedly with a model of the adversary using self play to determine the distribution of adversarial versus clean queries. It then strategically selects a classifier from a set of pre-trained classifiers that balances the likelihood of correct prediction for the query along with reducing the costs to use the classifier. We have evaluated our proposed technique using clean and adversarial text data with deep neural network-based classifiers and shown that the learner can select an appropriate classifier that is commensurate with the query type (clean or adversarial) while remaining aware of the cost to use the classifier.
\end{abstract}

\section{Introduction}
Adversarial machine learning~\cite{vorobeychik2018adversarial} is an important problem in machine learning based prediction systems such as email spam filters, online recommender systems, text classifier and sentiment analysis techniques used on social media, and, automatic video and image classifiers. The main problem in adversarial learning is to prevent an adversary from bypassing an ML-based predictive model such as  a classifier by sending engineered, malicious data instances called adversarial examples. These attacks, called evasion attacks, could enable a malicious adversary to subvert the learner's ML model and possibly get access to critical resources being protected by the learner. For instance, in the context of malware detection, an adversary could try to surreptitiously insert ill-formed PDF objects into a valid PDF file to convert it into a malware that could bypass a ML-based malware detector and subsequently crash an Internet browser attempting to read the corrupted PDF file. Researchers have proposed techniques including adversarial training~\cite{yuan2019adversarial} and game theory based techniques~\cite{Dasgupta19} to address the problem of adversarial learning. These techniques employ an approach called classifier hardening on a single classifier where the decision boundary of the classifier is refined over time via re-training with adversarial data. However, improving the robustness of single classifier hardening techniques is an open problem and these techniques are still been known to be susceptible to adversarial attacks~\cite{madry2017towards}. Moreover, classifier hardening techniques do not explicitly align costs to harden the classifier (e.g., costs to acquire adversarial training data, and, time and costs to harden it with adversarial data) with the data being classified. For instance, for classifying clean data, a classifier hardened over several batches of adversarial data might be excessive, as a classifier that is not hardened might achieve similar performance. In this paper, we posit that the costs of a classifier-based ML model to adversarial attacks can be improved without deteriorating classification accuracy, if instead of using a single classifier, we use multiple classifiers that are hardened separately against attacks of different strengths. Our idea is based on the well-known result of Wolpert's theorem~\cite{wolpert2002supervised} that there is not a single classifier which can be optimal for all classification tasks and multiple, combined classifiers could outperform the best individual classifier. The main challenge with using multiple classifiers is to determine the appropriate pairing between a query sent to the classifier, with either clean or adversarial data of different attacks strengths, and a commensurate classifier from the collection of classifiers to handle the query most effectively, e.g., with least likelihood of classification errors and while aligning classifier hardening costs with the query's attack strength. A further wrinkle to the problem is that the classifier is not aware whether the query is with clean data from a legitimate client versus with adversarial data from an attacker. To address this problem, we propose a game theoretic framework called a Repeated Bayesian Sequential Game with self play between a learner and an adversary. The outcome of the game strategically selects an appropriate classifier for the learner. Our proposed formulation enables us to realize several practical aspects of learner-attacker interactions including uncertainty of the learner about the strengths of different attacks, costs to the learner and attacker to train the classifier and generate adversarial examples respectively, rewards and penalties to attacker and learner for successes in their attacks and defenses respectively. Finally, a Bayesian game based representation enables our approach to handle asymmetric interactions between the learner and its clients for both non-competitive (legitimate clients, clean queries) and competitive (attackers, adversarial queries) settings. To the best of our knowledge, our work is one of the first attempts at using multiple classifiers deployed strategically to tackle the adversarial learning problem. We have validated our approach within a learner-adversary setting where the adversary generates queries with both clean and adversarial text data with different attack strengths while the learner's classifiers use deep network models for classification. Our results show that the learner can successfully converge to the distribution of different attacks types of the adversary and can strategically select different classifiers to reduce the overall classification cost without deteriorating the classification accuracy.


\section{Related Work}
\label{sec_relwork}
Early work in adversarial learning modeled the interaction between the learner and adversary as a competitive, 2-player game~\cite{dalvi2004adversarial}~\cite{globerson2006nightmare}. The game is solved as a constrained optimization problem and its solution provides an attack strategy for the attacker, e.g., which subset of features to modify in the data sent to the learner's classifier to effect an incorrect prediction by the classifier. While for the learner, a response strategy determines appropriate parameter values for its ML model, e.g., weights for regression or for a neural network model, so that the attack would be unsuccessful. Subsequently, researchers extended the adversarial learning game using different formalisation including a sequential game~\cite{Bruckner12}, a Bayesian game~\cite{Groshans13} where the learner has incomplete information about the attacker's strategy, a bi-level optimization problem ~\cite{Mei15}~\cite{Alfeld17}, strategic classification~\cite{dong2018strategic} and randomization over strategies of the learner and the adversary~\cite{Bulo17}. In most of these techniques, the learner's strategy at each instance of the game is to adjust its model parameters. While that might be practical for smaller models with few parameters, as the model size increases, e.g., for a deep network with thousands of parameters, the strategy might become infeasible to realize in practice. In contrast, in our work, we use the strategy output of the game to select an appropriate classifier for the learner from an existing, pre-trained set of classifiers. 

With the popularity of deep neural networks as ML models, several adversarial learning techniques based on adversarial training for deep networks have also been researched~\cite{goodfellow2014explaining}~\cite{kurakin2016adversarial}. In adversarial training, the learner's ML model is trained with both clean and adversarial data to improve its capability to correctly classify adversarial data. Unlike game theory representations, adversarial training techniques do not explicitly model costs, penalties and rewards for learner and attacker. In~\cite{tramer2017ensemble}, researchers have proposed an ensemble of ML models to generate adversarial examples and then use those examples from different models to harden a classifier. Most of these techniques use adversarial training to harden a single classifier. In contrast, instead of hardening one classifier, in our work, the learner maintains multiple classifiers with different degrees of hardening and strategically deploys one of them. Recently, researchers have proposed using ensembles of classifiers for adversarial training~\cite{bagnall2017training}~\cite{kariyappa2019improving}~\cite{li2018enhancing} that use a diversity measure between classifier ensembles to determine if an instance is adversarial versus clean. Generative Adversarial Nets (GANs)~\cite{goodfellow2014generative} also model interaction between an adversary (generator) and learner (discriminator) as a $2$-player game. However, the objectives of GANs and adversarial learning are different. GANs enable an adversary to refine its data generation process, starting from a random distribution, so that the generated data is indistinguishable from legitimate data. Adversarial learning, on the other hand, aims to enable the learner to strategically defend against adversarial attacks. Recently, security games~\cite{Tambe11} and adversarial games for network security have been proposed in~\cite{schlenker2018deceiving}. The proposed technique employs deception by the learner to misguide the adversary, which can be considered as a complimentary approach to the techniques proposed here, for building defenses against adversarial attacks.

\section{Adversarial Learning as  Bayesian Game}
\label{sec_problem}
We consider a supervised learning setting for binary classification where learner, ${\cal L}$, receives data instances as queries from an attacker or adversary, ${\cal A}$. We represent this interaction between ${\cal L}$ and ${\cal A}$ as a $2$-player Bayesian game for adversarial learning~\cite{Groshans13} while adapting Gro{\ss}hans' model to multiple classifiers, different attack strengths and repeated interactions between ${\cal L}$ and ${\cal A}$. We describe the different components of the game below:

Let $\mathbf{X}^{ev}$ denote a set of queries. We refer to $\mathbf{X}^{ev}$ as the clean query set. Let $ X = (\mathbf{x}, y), X \in \mathbf{X}^{ev}$ denote a query data instance, where $\mathbf{x} = \{x_1, x_2, ...\}$ is its set of attributes or features and $y \in \{0,1\}$ is its ground truth label. 

{\em Adversary.} ${\cal A}$ sends either clean or adversarial data as queries; the latter is generated by perturbing clean data using a perturbation function $\delta: \mathbf{x} \rightarrow \mathbf{x}$. We assume that ${\cal A}$ uses different perturbation functions $\delta_i, \, i=0, 1, 2,...$, where $i$ denotes the strength of the perturbation. For example, perturbation strength could correspond to the number of features of $\mathbf{x}$ that are modified to convert it into an adversarial instance~\cite{globerson2006nightmare}. $\delta_i(\mathbf{x})$ denotes the adversarial data generated with perturbation strength $i$ and $\delta_{i+1}$ is a stronger perturbation than $\delta_i$. Perturbing $\mathbf{x}$ does not change its ground truth label, $y$. For notational convenience, we refer to clean data, $\mathbf{x} = \delta_0(\mathbf{x})$. An action for ${\cal A}$ is to select a $\delta_i$, use it to convert clean instance $\mathbf{x}$  into adversarial instance $\delta_i(\mathbf{x})$, and send the adversarial instance to ${\cal L}$. 

{\em Learner.} ${\cal L}$ receives a query data instance $\mathbf{\bar{x}}$ and its task is to correctly predict its category. ${\cal L}$ is neither aware of the perturbation strength $i$ of $\mathbf{x}^{\delta_i}$ inside the data, nor is it aware of $\bar{y}$, the ground truth label of $\mathbf{\bar{x}}$. ${\cal L}$ uses a set of classifiers, $L_j,\, j=0, 1, 2...$ for its prediction task. $L_j$ implements a classification, $L_j: \mathbf{x} \rightarrow \{0, 1\}$, that outputs a category given the features of the query data. Classifier $L_j$ is adversarially trained using training data $\mathbf{X}^{tr, \delta_j} \notin \mathbf{X}^{ev}$, where $\delta_j$ denotes the perturbation strength of the training data. We assume that $L_{j+1}$ is a stronger classifier than $L_j$: for a query $\mathbf{x}$, $L_{j+1}$ has a higher confidence in its output than $L_j$, or, mathematcally, $P(L_{j+1}(\mathbf{x}) = y) \geq P(L_j(\mathbf{x}) = y)$. An action for ${\cal L}$  is to select a classifier $L_j$ and use it to classify the data instance sent by ${\cal A}$. We denote the action set of ${\cal L}$ as $Ac_{\cal L} = \{L_0, L_1, L_2...\}$. Let $\Pi(Ac_{\cal L})$ be the set of probability distributions over $Ac_{\cal L}$. $s_{\cal L} \in \Pi(Ac_{\cal L})$ denotes a strategy for ${\cal L}$ and $s_{\cal L}(L_j)$the probability of selecting $L_j$ under strategy  $s_{\cal L}$. Finally, recall that ${\cal L}$ is not aware of the perturbation $\delta_i$ that has been used by ${\cal A}$ on the query data instance, $\bar{X}$ that it receives. To model this uncertainty about its opponent, ${\cal L}$ uses epistemic types for ${\cal A}$~\cite{harsanyi1967games}. ${\cal A}$'s type $\theta_i$ denotes that ${\cal A}$ uses perturbation strength $i$ to create $\mathbf{\bar{x}}$, i.e., $\mathbf{\bar{x}} = \mathbf{x}_{\theta_i} = \delta_i(\mathbf{x})$. $\Theta_{\cal A} = \{\theta_i\}$ is ${\cal A}$'s set of types and $p: \Theta_{\cal A} \rightarrow [0, 1]^{|\theta_{\cal A}|}$ denotes a probability distribution over these types. $\Theta_{\cal A}$ is known to ${\cal L}$, $p()$ is calculated by ${\cal L}$. But $\theta_i$, the exact realization of ${\cal A}$'s type (in other words, the perturbation strength used to create $\mathbf{\bar{x}}$) is not known to ${\cal L}$ when it receives $\mathbf{\bar{x}}$ from ${\cal A}$.

{\em Utilities.} Utilities are numeric values assigned by each player to the outcomes from the players' joint actions in a game. Each player could then preferentially rank its joint outcomes and select a suitable action such as a utility maximizing action. For our game, recall that ${\cal L}$ is not able to observe ${\cal A}$'s type $\theta_i$ (amount of perturbation in $\mathbf{\bar{x}}$). Therefore, ${\cal L}$ calculates an expected utility over ${\cal A}$'s possible types, $\Theta_{\cal A}$, using ${\cal A}$'s type distribution $p()$. ${\cal L}$'s expected utility for strategy $s_{\cal L}$ with query data $\mathbf{\bar{x}}$ and ground truth label $\bar{y}$ is given by:
\begin{multline}
\label{eqn_UL}
EU_{\cal L}(s_{\cal L}, \mathbf{\bar{x}}, \theta_{\cal A}, p())= \sum_{\theta_i \in \Theta_{\cal A}} p(\theta_i)  U_{\cal L}(L_j, \mathbf{\bar{x}}, \theta_i),\\
U_{\cal L}(L_j, \mathbf{\bar{x}}, \theta_i) = \sum_{L_j} s_{\cal L}(L_j)\Big(P(L_j(({\mathbf{\bar{x}}}) = \bar{y})|\theta_i)v_{\cal_L}(L_j, \theta_i) \\
- c_{L_j}\Big), 
\end{multline}
\noindent where $P((L_j({\mathbf{\bar{x}}}) = \bar{y})|\theta_i)$ is the probability that ${\cal L}$ makes a correct prediction given $\mathbf{\bar{x}}$ was generated using $\theta_i$, $v_{\cal L}(L_j, \theta_i)$ is the value for ${\cal L}$ from classifying $\mathbf{\bar{x}}$ using $L_j$ and $c_{L_j}$ is the cost of using classifier $L_j$.

In adversarial settings, it is usually assumed that the adversary is aware of the learner's  prediction model, e.g., model parameters of the learner's classifier~\cite{Alfeld17}. In the context of our game, this can be interpreted as ${\cal A}$ knowing ${\cal L}$'s strategy, $s_{\cal L}$. ${\cal A}$'s utility for query data $\mathbf{\bar{x}}$ with ground truth label $\bar{y}$, for ${\cal L}$'s strategy $s_{\cal L}$ and its own type $\theta_i$ is given by:
\begin{multline}
\label{eqn_UA}
U_{\cal A}(s_{\cal L}, \mathbf{\bar{x}}, \theta_i) = \sum_{L_j} s_{\cal L}(L_j) \Big(P(L_j({\mathbf{{\bar{x}}}}) \neq \bar{y})v_{\cal_A}(L_j, \theta_i) \\
- c_{\theta_i}\Big), 
\end{multline}

\noindent where $P(L_j(\mathbf{\bar{x}}) \neq \bar{y})$ represents the probability that ${\cal L}$ makes a mistake in prediction (in other words, ${\cal A}$'s adversarial perturbation of clean data was successful) and $v_{\cal A}(L_j, \theta_i)$ is the value that ${\cal A}$ derives from sending the query data $\mathbf{x}_{\theta_i}$ when ${\cal L}$'s action is $L_j$ and $c_{\theta_i}$ is ${\cal A}$'s cost for generating adversarial data with type (perturbation strength) $\theta_i$.

{\em Bayesian Sequential Game.} Using the above actions and utility functions, we can represent a Bayesian sequential game between ${\cal L}$ and ${\cal A}$ as $\Gamma = [N, Ac, U, \Theta_A, p()]$, where $N=\{{\cal L}, {\cal A}\}$ is the set of players, $Ac = Ac_{\cal L} \times \Theta_{\cal A}$ is the set of joint action-types of ${\cal L}$ and ${\cal A}$, $U= (EU_{\cal L}, U_{\cal A})$ denotes the utilities received by ${\cal L}$ and ${\cal A}$ (given in Eqns. \ref{eqn_UL} and \ref{eqn_UA}), $\Theta_{\cal A}$ and $p()$ are the set of ${\cal A}$'s types and probability distribution over those types, as defined before. 

The computational problem facing ${\cal L}$ and ${\cal A}$ is to calculate a suitable strategy $s^*_{\cal L}$ and suitable type $\theta^*_i$ respectively. To do this calculation using Eqn. \ref{eqn_UL}, ${\cal L}$ also needs to know the value of $p()$, the probability distribution over ${\cal A}$'s types.  To address these issues, we propose an approach using a technique called self play  with repeated plays of the above Bayesian Sequential game called a Repeated Bayesian Sequential Game (RBSG), as described below.

\subsection{Repeated Bayesian Sequential Game and Self-Play}
\label{sec_rbsg}
The objective of ${\cal L}$ is to determine a suitable strategy $s_{\cal L}^*$ to play against ${\cal A}$ that would improve its expected utility by deploying an appropriate classifier that has been hardened commensurate to the strength of the perturbation used by ${\cal A}$. 
To achieve this, ${\cal L}$ uses self play, where ${\cal L}$ and ${\cal A}$ play the Bayesian Sequential game, $\Gamma$, repeatedly. For the sake of legibility, we continue to use the notation ${\cal A}$ to denote ${\cal L}$'s self play adversary. The repeated interactions between ${\cal L}$ and ${\cal A}$ can be represented as a game tree with sequential moves between them. A node in the game tree denotes a player's turn to make a move. In a move, a player selects an action from its action set. ${\cal L}$ and ${\cal A}$ make alternate moves with ${\cal L}$ moving first. A pair of moves by ${\cal L}$ and ${\cal A}$ corresponds to an instance of the Bayesian Sequential game, $\Gamma$, realized as below.

\begin{algorithm}[htb!]
\label{algo_game_play}
	\caption{game-play()}
    {
	    Select $s^*_{\cal L}$ using current belief of $\hat{p}$, and $\theta^*_i$ (Eqn.~\ref{eqn_NE} or ~\ref{eqn_UCB})\\ 
		Calculate utils. recd.: $\hat{u}_{\cal L}$ and $\hat{u}_{\cal A}$ with observed values of $s^*_{\cal L}$ and $\theta_i^*$ resp. (using Eqns.~\ref{eqn_UL} and~\ref{eqn_UA}) \\
		return $ (\hat{u}_{\cal L}, \hat{u}_{\cal A})$
	}
\end{algorithm}

{\em Game Play.} As shown in Algo. ~\ref{algo_game_play}, ${\cal L}$'s moves by selecting a strategy $s^*_{\cal L}$. ${\cal A}$ then selects type (perturbation strength) $\theta_i^* \sim p()$ while observing $s^*_{\cal L}$. With the selected $\theta_i^*$, ${\cal A}$ then generates $q$ adversarial queries by perturbing $q$ clean data instances from $\mathbf{X}^{ev}$, and sends each adversarial query, $\mathbf{\bar{x}}$, to ${\cal L}$. After ${\cal L}$ processes the queries, both ${\cal L}$ and ${\cal A}$ receive utilities given by Eqns. \ref{eqn_UL} and \ref{eqn_UA} respectively.  The problem facing ${\cal L}$ is to calculate $s^*_{\cal L}$ without observing  $\theta_i^*$ and $p()$ from ${\cal A}$'s moves. We solve this problem using a modified Monte Carlo Tree Search (MCTS) algorithm, as described below. 

\subsubsection{Calculating strategy $s_{\cal L}^*$.}
To calculate $s_{\cal L}^*$, ${\cal L}$ generates different paths in the game tree to discover utilities received from different sequences of moves. To systematically explore the game tree, ${\cal L}$ uses an MCTS-like algorithm~\cite{browne2012survey}, called {\em TreeTraverse}. shown in Algos. ~\ref{algo_TreeTraverse} and ~\ref{algo_rollout}. {\em TreeTraverse} works by generating a sequences of moves  or game plays corresponding to a path in the game tree up to a finite cutoff depth $h$. ${\cal L}$ and ${\cal A}$'s utilities from their moves are recorded along the path and once the bottommost level is reached, the utilities are updated along the path upwards toward the root. In this way, moves that could lead to high utility can be identified by each player. 

The key aspects of MCTS are to balance exploration and exploitation while traversing the game tree by using a heuristic function called {\em selectBestChild} (Algo. ~\ref{algo_TreeTraverse}, line $4$), and, doing an operation called rollout to rapidly traverse unexplored parts of the game tree by selecting actions for each player up to the game tree's cutoff depth $h$ (Algo. ~\ref{algo_rollout}). In our {\em TreeTraverse} algorithm, we have used two heuristic functions for {\em selectBestChild}, as described below:

{\em Bayes Nash Equilibrium (BNE).} In BNE, each player selects a best reponse strategy that maximizes its utilities, given the possible strategies of its opponent~\cite{harsanyi1967games}. The strategies for ${\cal L}$ and ${\cal A}$ calculated using BNE are given by:
\begin{flalign}
s_{\cal L}^* = \displaystyle arg \,\max_{s_{\cal L} \in \Pi(Ac_{\cal L})} EU_{\cal L}(s_{\cal L}, \mathbf{\bar{x}}, \Theta_{\cal A}, p()), \nonumber \\
\theta^*_i =  \displaystyle arg\,\max_{\theta_i \in \Theta_{\cal A}} \; U_{\cal A}(s_{\cal L}^*, \mathbf{\bar{x}}, \theta_i), 
\label{eqn_NE}
\end{flalign}
\noindent where $u_{\cal A}$ is given by Eqn.~\ref{eqn_UA} and $EU_{\cal L}$ is given by Eqn.~\ref{eqn_UL} with ${\cal A}$'s actual type distribution $p(\theta_i)$ replaced by ${\cal L}$'s belief distribution $\hat{p}(\theta_i)$.

{\em Upper Confidence Bound (UCB).} UCB is a bandit-based technique~\cite{browne2012survey} that weighs the expected utility of a move with the number of times it has been visited, so that previously unexplored or less-explored actions at a move are also tried. UCB  uses the following equation to calculate $s_{\cal L}^*$ and $\theta_i^*$:
\begin{eqnarray}
{\scriptstyle
s_{\cal L}^* = \arg \max_{\Pi(L_j)}  \sum_{\theta_i} \left (p(\theta_i) \sum_{\mathbf{\bar{x}} \in \mathbf{\bar{X}}} U_{\cal L}(L_j, \mathbf{\bar{x}}, \theta_i) + C \sqrt{\frac{2\ln{Par_{visit}}}{L_{j, visit}}} \right)
}\nonumber \\
{\scriptstyle 
\theta_i^* = \arg \max_{\theta_i} \sum_{L_j} \left( \sum_{\mathbf{\bar{x}} \in \mathbf{\bar{X}}} s_{\cal L}^*(L_j) U_{\cal A}(L_j, \mathbf{\bar{x}}, \theta_i) + C \sqrt{\frac{2\ln{Par_{visit}}}{\theta_{i, visit}}} \right)
}
\label{eqn_UCB}
\end{eqnarray}

Here, $C$ is a constant, $Par_{visit}$ is the number of times the parent node of the current node was visited and $L_{j, visit}$ and $\theta_{i, visit}$ are the number of times the current node has been visited for ${\cal L}$ and ${\cal A}$ respectively.

\begin{algorithm}[htb!]
\label{algo_TreeTraverse}
	\caption{TreeTraverse($v$)}
	\KwIn{$v$: start node for traversal}
	\KwOut{$v_{val}$: value from tree traversal (via backtracking) starting from $v$ up to depth $h$}
    \If {$v_{depth} = h$}
	{
	    return
	}
	{\bf else} \If{$v$ is fully expanded}
    {
        $c_{val} \leftarrow$ TreeTraverse(selectBestChild($v$)) // go down game tree along best action (Eqns. \ref{eqn_NE} or \ref{eqn_UCB})\\
        Update $v_{val} \leftarrow v_{val} + c_{val}$; increment $v_{visit}$\\
        return $c_{val}$
    }
    {\bf else} \If {$v$ is visited but not expanded}
    {
        $\bar{c} \leftarrow$ generatedAllChildren($v$) // all actions\\
        $c \leftarrow$ select random child (action) from $\bar{c}$\\
        $c_{val} \leftarrow$ rollout($c$)\\
        Update $v_{val} \leftarrow v_{val} + c_{val}$, 
        increment $v_{visit}$ and $c_{visit}$\\
        return $c_{val}$
    }
    {\bf else} \If {$v$ is not visited}
    {
        $v_{val} \leftarrow$ rollout($v$)\\
        Increment $v_{visit}$\\
        return $v_{val}$
    }
\end{algorithm}

\begin{algorithm}[htb!]
\label{algo_rollout}
	\caption{Rollout($v$)}
	\KwIn{$v$: start node for rollout}
	\KwOut{$v_{val}$: value from rollout (via backtracking) starting from $v$ up to depth $h$}
    \eIf {$v$ is terminal}
	{
	    $\hat{u}_{\cal L}, \hat{u}_{\cal A} \leftarrow$ game-play() \\
		return $ (\hat{u}_{\cal L}, \hat{u}_{\cal A})$
	}
    {
        $c \leftarrow$ select child of $v$ prop.
        to $u_{\cal L}$ (for ${\cal L}$'s move) or 
        prop. to $p()$ (for ${\cal A}$'s move)\\
        $c_{val} \leftarrow$ rollout($c$)\\
        return $c_{val}$
    }
\end{algorithm}

\subsubsection{Updating belief of ${\cal A}$'s type distribution.} The {\em TreeTraveese} algorithm explores a sequence of moves along aany single path from the root of the game tree up to the cutoff depth $h$. We call this a trial for the RBSG. To update its belief distribution $\hat{p}$, ${\cal L}$ uses multiple trials and, at the end of each trial, ${\cal L}$ uses an update strategy to update $\hat{p}()$. We consider two probability update strategies that can be used by ${\cal L}$ (Algo.~\ref{algo_self_play}, line 4) for updating $p_{\tilde{\Theta_{\cal A}}}$. {\bf 1) Fictitious Play (FP)}: In fictitious play~\cite{Leytonbrown09}, the probability of type $\theta_i$ is the fraction of times it was played following action $L_j$, as given by the following update rule:
\begin{equation}
    P({\theta_i}|L_j) = \frac{\mbox{No. of times}\, \theta_i\,\mbox{selected after}\,L_j}{\mbox{Total no. of times}\,L_j\,\mbox{selected}}
    \label{eqn_FP}
\end{equation} 
{\bf 2) Bayesian Update (BU)}: Bayesian update of $\theta_i$ calculates the conditional probability of selecting $\theta_i$ when it followed $L_j$ using Bayes rule, given by the following equation:
\begin{equation}
    P({\theta_i}|L_j) = \frac{P(L_j|\theta_i)P(\theta_i)}{P(L_j)}
= \frac{P(L_j|\theta_i)P(\theta_i)}{\sum_{\theta_i} P(L_j|\theta_i)P(\theta_i)},
\label{eqn_BU}
\end{equation}
where $P(L_j|\theta_i)$ is the fraction of times $L_j$ was played following $\theta_i$, $P(L_j)$ is known to ${\cal L}$ and the denominator is a normalization term. The updated probability estimate is then used by ${\cal L}$ to calculate the expected utilities in Eqns.~\ref{eqn_NE} and ~\ref{eqn_UCB} for its actions more accurately against ${\cal A}$'s in future trials. 

\begin{algorithm}[htb!]
\label{algo_self_play}

	\caption{Self-Play()}
	\For {$\tau = 1...n_{trials}$}
	{
	    $root \leftarrow$ ${\cal L}$'s first move with randomly sel. action\\
		TreeTraverse(root)\\
	    Update $\hat{p}$ using prob. update strategy (fic. play, Eqn.~\ref{eqn_FP} or Bayes update, Eqn.~\ref{eqn_BU})\\
    }
\end{algorithm}

\section{Experimental Results}
\label{sec_expt}
We have evaluated the performance our proposed RBSG with self play-based adversarial learning technique for a binary classification task with text data using the Yelp review polarity data set. ~\cite{datasets}. Each data instance has either of two labels, $1$ (negative) and $2$ (positive). The clean training and test sets have $560,000$ and $38,000$ samples respectively. We used the Character Convolutional Neural Network (CharCNN)~\cite{zhang2015character}  model that consists of $5$ convolution layers followed by $3$ fully connected layers. It uses convolution layers to identify character level features to classify text. For generating adversarial text, we used the single character gradient based replacement technique~\cite{Liang18}. Given a data instance in the form of a text character string as input to an ML model, the method works by classifying the text using the model and calculating the gradient of the loss function for each character in the input text. It then replaces the character with the most negative gradient (most influential on the classifier output) in the text with the character that has the least positive gradient (least influential on the classifier output). The technique can be used iteratively on a data instance to replace multiple characters in the text and create adversarial text with different attack strengths, e.g., two iterations of the technique yields adversarial text with perturbation strength $2$. All experiments were performed on a computer with $20$ dual core, $2.3$\,GHz Intel Xeon CPUs with Nvidia Tesla K40C GPU. The RBSG self play code was implemented in Python $2.7$; the CharCNN and adversarial text generataion code used Tensorflow $1.11$ for building and training their deep network models. The CharCNN was first trained with clean data, and then hardened separately with two adversarial training data sets with $200,000$ adversarial training samples of perturbation strengths $1$ and $2$ respectively. This gave three classifiers for ${\cal L}$ with increasing hardening levels, denoted by $ L_0, L_1$ and $L_2$. The accuracies of these classifiers were then evaluated with $50,000$ instances of test data of perturbation strengths $1$, $2$ and $3$ each, as reported in Table~\ref{table_classifier_acc}.

\begin{table}[]
    \centering
    \begin{tabular}{|c|c|c|c|}
    \hline
                & $L_0$    & $L_1$    & $L_2$  \\
    \hline
        Clean   & $0.9392$ & $0.9426$ & $0.94$         \\
    \hline
        Adv $1$ & $0.8684$& $0.88$   & $0.8782$       \\
    \hline
        Adv $2$ & $0.7706$ & $0.7922$ & $0.8152$       \\
    \hline
        Adv $3$ & $0.6814$ & $0.7056$ & $0.7502$       \\
    \hline
    \end{tabular}
    \caption{Testing accuracy of individual classifiers with different hardening levels (columns) on adversarial test data with different perturbation strengths (rows).}
    \label{table_classifier_acc}
\end{table}

Adversary ${\cal A}$ generates queries with either clean data or adversarial data with perturbation strengths $1, 2$ and $3$, giving $\Theta_{\cal A} = \{\theta_0, \theta_1, \theta_2, \theta_3\}$. ${\cal L}$ uses three classifiers, so, $Ac_{\cal L} = \{L_0, L_1, L_2\}$. The different parameters used for our experiments are: cutoff depth in self play, $h=20$; number of trials in self play, $n_{trials}=10$; batch size for queries sent by ${\cal A}$ to ${\cal L}$, $q=10$; and constant in UCB calculation (Eqn.~\ref{eqn_UCB}), $C=2$.

\begin{table}[htb!]
    \centering
    \begin{tabular}{|c|c|c|c|c|}
    \hline
    {\bf UCB}   & $L_0$    & $L_1$    & $L_2$    & Acc. \\
    \hline
        Clean   & $43.75\%$ & $29.46\%$ & $26.79\%$   & $0.9321$ \\
    \hline
        Adv $1$ & $39.65\%$ & $24.13\%$ & $36.21$   & $0.8716$ \\
    \hline
        Adv $2$ & $24.11\%$ & $25\%$ & $50.89\%$   & $0.8062$ \\
    \hline
        Adv $3$ & $39.81\%$ & $20.37\%$ & $39.81\%$   & $0.7222$ \\
    \hline
    \end{tabular}
    
    \begin{tabular}{|c|c|c|c|c|}
    \hline
    {\bf BNE}   & $L_0$    & $L_1$    & $L_2$    & Acc. \\
    \hline
        Clean   & $57.56\%$ & $10.37\%$ & $32.07\%$   & $0.9302$ \\
    \hline
        Adv $1$ & $33.91\%$ & $46.96\%$ & $19.13$   & $0.867$ \\
    \hline
        Adv $2$ & $29.46\%$ & $27.68\%$ & $42.86\%$   & $0.808$ \\
    \hline
        Adv $3$ & $31.53\%$ & $32.43\%$ & $36.04\%$   & $0.709$ \\
    \hline
    \end{tabular}
    \caption{Percentage of different classifiers used and accuracies (columns) obtained for clean and adversarial data of different perturbation strengths (rows). Data in the top and bottom tables are with Upper Confidence Bound (UCB) and Bayes Nash Equilibrium (BNE), respectively, for action selection during self play. }
    \label{table_ratios_UCB_NE}
\end{table}

\begin{figure}[htb!]
    \centering
    \includegraphics[width=3.2in]{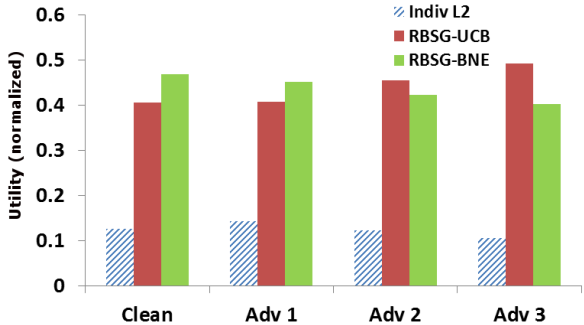}
    \caption{Relative utilities obtained by individual classifier $L_2$, and RBSG with self play-based techniques with UCB and BNE action selection for data reported in Table~\ref{table_ratios_UCB_NE}}.
    \label{fig_utils}
\end{figure}

For our first set of experiments, we validated if ${\cal L}$, using the self play algorithm, could effectively deploy appropriate classifiers for data of different perturbation strengths. We created four different type distributions for data generated by ${\cal A}$, each distribution having $98\%$ of one of the four types$\{\theta_0, \theta_1, \theta_2, \theta_3\}$. ${\cal L}$ used either Upper Confidence Bound (UCB) or Bayes Nash Equilibrium (Eqn.~\ref{eqn_NE} or Eqn.~\ref{eqn_UCB}) to select actions in the game tree during self play. Our results are shown in Table ~\ref{table_ratios_UCB_NE}. The results show that both UCB and BNE metric for action selection perform comparably. The accuracy obtained using our RBSG-based self play technique on clean and adversarial data perturbed with different perturbation strengths (last column of Table~\ref{table_ratios_UCB_NE} is not degraded and comparable to the best accuracies obtained with the most hardened classifier, $L_2$, when used individually (column $4$ of Table~\ref{table_classifier_acc}). The RBSG with self play technique is also able to align adversarial data of different perturbation strengths with the commensurately hardened classifier, as shown by the maximum percentage of each row in Table~\ref{table_classifier_acc} corresponding to the classifier hardened with adversarial data of that perturbation strength. Note that with adversarial data of perturbation strength $3$, $Adv\,3$, the classifiers are selected almost uniformly. This is because none of the classifiers, $L_0, L_1$ or $L_2$ were trained with adversarial data of perturbation strength $3$. $L_2$, which had the highest individual accuracy for $Adv\,3$ data, is used most frequently, albeit marginally, for $Adv\,3$ data in Table~\ref{table_ratios_UCB_NE}. Our Self-play technique also strategically also uses $L_0$ and $L_1$ that incur lower costs to deploy than $L_2$. Consequently, the utility obtained  by ${\cal L}$ with self play is better than its utility while using individual classifier $L_2$ only. Fig.~\ref{fig_utils} shows the comparison of the relative utilities obtained by ${\cal L}$ while using the proposed RBSG with self play technique versus the utilities obtained while using the most hardened individual classifier $L_2$. As illustrated, the RBSG with self play technique is able to improve utilities as it deploys lower cost classifiers $L_0$ and $L_1$ along with $L_2$ while aligning the expected perturbation strength of the query data, estimated via $\hat{p}$, with the commensurately hardened classifier.

\begin{figure}
    \centering
    \includegraphics[width=3.2in]{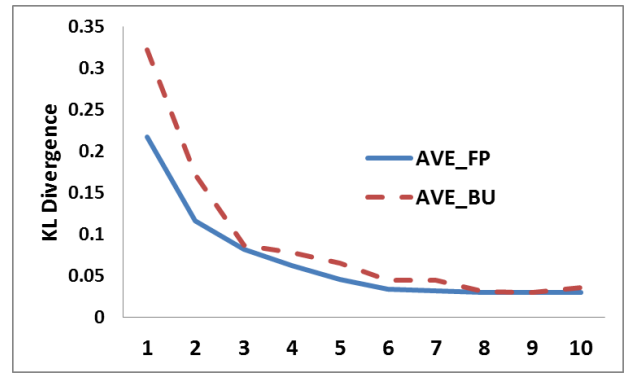}
    \caption{KL divergence  between ${\cal A}$'s actual type distribution and ${\cal L}$'s belief distribution using fictitious play and Bayesian update for $n_{trials} = 10, h=20$. Results are averaged over $10$ runs.}
    \label{fig_kldiv}
\end{figure}
For our next experiments, we evaluated the convergence of ${\cal L}$'s belief distribution $\hat{p}()$ to ${\cal A}$'s actual type distribution $p()$ using the fictitious play and Bayesian update probability update strategies (Eqns.~\ref{eqn_FP} and ~\ref{eqn_BU}).  Results were averaged over $10$ runs. For each run, $p()$ was selected as a random distribution. We report the Kullback-Liebler(KL) divergence between $\hat{p}()$ and $p()$, given by $D_{KL}(\hat{p}||p) = \sum_{\theta_i \in \Theta_{\cal A}} \hat{p}(\theta_i)ln\frac{\hat{p}(\theta_i)}{p(\theta_i)}$. As shown in Fig.~\ref{fig_kldiv}, with both strategies $\hat{p}$ is able to converge to within $5\%$ of $p()$ within about $6$ trials. Fictitious play converges faster with higher KL divergence values while Bayesian update takes a longer time to converge owing to its more complex calculations.

\section{Conclusion}
\label{sec_conclusion}
We proposed a technique for improving the costs of a classifier-based ML model against adversarial attacks of different strengths without deteriorating its performance by using repeated game-like interactions between a learner and an adversary. There are several important directions that are worthy of further investigation. First, the assumption in existing research which assumes that the learner reveals its classifier to the adversary is rather limiting. A more realistic situation would be that the adversary is able to reverse engineer the learner's classifiers, but it is not aware of the frequency with which the learner deploys them. The adversary could then also build a model of the learner via repeated interactions to determine its perturbation strength strategically. Secondly, although used as a popular solution technique in games, Nash equilibrium (NE) strategy calculation is known to have certain shortcomings such as assuming that players always behave rationally. In reality, an adversary could behave myopically, select a greedy outcome, or, adopt sub-optimal, low and slow strategies to misguide the learner. To handle these situations, a direction we are interested in exploring is to use recent techniques such as regret-based techniques, safety value and exploitability of opponents, instead of Bayes Nash equilibrium-based strategy selection. Finally, integrating reinforcement learning for our adversarial learning setting promises to be another direction worthy of further investigation.

{\small 
\bibliographystyle{aaai}
\bibliography{refs}
}
\end{document}